\documentclass[letterpaper, 10 pt, conference]{ieeeconf}  

\IEEEoverridecommandlockouts                              

\usepackage[noadjust]{cite}

\usepackage{graphicx} 
\usepackage{comment}
\usepackage{booktabs} 
\usepackage{xcolor}

\newcommand\blue[1]{\textcolor{blue}{#1}}
\usepackage{bm} 
\usepackage{amsmath}
\usepackage{amssymb} 
\usepackage{mathrsfs} 
\usepackage{algorithm} 
\usepackage{mathtools} 

\newtheorem{lemma}{Lemma}
\newtheorem{theorem}{Theorem}
\let\labelindent\relax 
\usepackage{enumitem}
\usepackage{cuted} 
\usepackage{siunitx}

\usepackage{ulem}

\usepackage{kotex}

\usepackage{url}
\usepackage[font=small]{caption}
\usepackage{subcaption}

\newcommand{\soutMath}[1]{\ifmmode\text{\sout{\ensuremath{#1}}}\else\sout{#1}\fi}

\newcommand\Tanh{\text{Tanh}}
\newcommand\Cosh{\text{Cosh}}
\newcommand\sgn{\text{sgn}}
\newcommand\norm[1]{\lVert #1 \rVert}

\usepackage{tikz}

\newcommand\copyrighttext{%
  \footnotesize \textcopyright 2024 IEEE.  Personal use of this material is permitted.  Permission from IEEE must be obtained for all other uses, in any current or future media, including reprinting/republishing this material for advertising or promotional purposes, creating new collective works, for resale or redistribution to servers or lists, or reuse of any copyrighted component of this work in other works.}
\newcommand\copyrightnotice{%
\begin{tikzpicture}[remember picture,overlay]
\node[anchor=south,yshift=10pt] at (current page.south) {\fbox{\parbox{\dimexpr\textwidth-\fboxsep-\fboxrule\relax}{\copyrighttext}}};
\end{tikzpicture}%
}

\title{\LARGE \bf
Saturated RISE control for considering rotor thrust saturation\\of fully actuated multirotor
}

\author{Dongjae Lee, H. Jin Kim
\thanks{This work was supported by Unmanned Vehicles Core Technology Research and Development Program through the National Research Foundation of Korea(NRF) and Unmanned Vehicle Advanced Research Center(UVARC) funded by the Ministry of Science and ICT(NRF-2020M3C1C1A010864).}%
\thanks{Dongjae Lee, H. Jin Kim are with the Department of Aerospace Engineering, Seoul National University (SNU), Seoul 08826, South Korea {\tt\small \{ehdwo713, hjinkim\}@snu.ac.kr}}%
}

\begin{document}

\maketitle
\copyrightnotice
\thispagestyle{empty}
\pagestyle{empty}

\begin{abstract}
This work proposes a saturated robust controller for a fully actuated multirotor that takes disturbance rejection and rotor thrust saturation into account. A disturbance rejection controller is required to prevent performance degradation in the presence of parametric uncertainty and external disturbance. Furthermore, rotor saturation should be properly addressed in a controller to avoid performance degradation or even instability due to a gap between the commanded input and the actual input during saturation. To address these issues, we present a modified saturated RISE (Robust Integral of the Sign of the Error) control method. The proposed modified saturated RISE controller is developed for expansion to a system with a non-diagonal, state-dependent input matrix. Next, we present reformulation of the system dynamics of a fully actuated multirotor, and apply the control law to the system. The proposed method is validated in simulation where the proposed controller outperforms the existing one thanks to the capability of handling the input matrix. 
\end{abstract}

\section{Introduction}

Uncertainty in physical parameters or external disturbance like wind degrades the control performance of a multirotor if not properly considered.
To address this issue, various disturbance rejection controllers have been developed using disturbance observer, sliding mode control, or backstepping control \cite{lee2021aerial, flores2023robust, ricardo2022smooth, chebbi2022robust, lee2022rise}. However, none of these works consider physical limit of actuators, i.e. rotors. If rotor saturation is not taken into account in a control law, a gap may exist between the input command computed from a controller and the actual input generated from an actuator, and this may lead to performance degradation.

To handle such rotor saturation problem, there have been research manipulating force and torque generated from a multirotor. A priority-based control strategy to relax the total thrust and the body torque is presented for a conventional multirotor in \cite{faessler2017thrust}, and \cite{franchi2018full} suggests a controller saturating body force in $x,y$ directions for a fully actuated multirotor. However, stability during the proposed relaxation strategy is not analyzed in the former, and the latter work considers saturation only in the force level and not in the rotor level.

Owing to the actuation principle of a multirotor, cross-coupling exists in a mapping from rotor thrusts to force and torque (e.g. for a fully actuated multirotor considered in this work, such mapping is represented by a matrix in (\ref{eq: matrix A})). Accordingly, a box constraint on the rotor thrust $u_i \in [\underline{u}_i,\overline{u}_i]$ does not imply a box constraint on a force/torque level. Thus, if to consider a box constraint on a force/torque level as in \cite{shao2022adaptive,xie2022saturated}, the box constraint should be defined in a conservative manner not to exceed the rotor thrust saturation limit. For this reason, \cite{wehbeh2022mpc, ghignoni2021anti, convens2017control} consider saturation directly in the rotor thrust level. However, only system dynamics without uncertainty and disturbance is addressed, and they lack disturbance rejection. 

This work presents a saturated RISE (Robust Integral of the Sign of the Error) controller for a fully actuated multirotor that directly considers saturation in the rotor thrust level and enables disturbance rejection. The proposed controller is motivated by \cite{fischer2014saturated, fischer2012saturated}, and we further develop the control law to be applicable to a fully actuated multirotor. The existing works consider a system dynamics in the form of $M(q) \ddot{q} = f(q,\dot{q},t) + \tau + d(t)$ where $M(q)$ is a symmetric, positive-definite matrix, and design a controller for $\tau$. However, if $u$ is the control input to be designed for $\tau = B(q)u$ with $B(q)$ being a non-diagonal, time-varying input matrix, then the existing method cannot be directly applied due to the input matrix $B(q)$. In this respect, this work presents a modified saturated RISE control for a system with a non-diagonal, state-dependent input matrix, and derives a rotor saturation-aware robust control for a fully actuated multirotor.

The contribution of this work can be summarized as follows:
\begin{itemize}
    \item We propose a controller for a fully actuated multirotor that considers actuator (rotor) level saturation and ensures disturbance rejection.
    \item We present a modified version of the saturated RISE control \cite{fischer2014saturated} to additionally handle a non-diagonal, state-dependent input matrix and apply the method to a fully actuated multirotor.
\end{itemize}

\subsection{Notations and preliminaries}
For a vector $v$, $v_i$ indicates the $i^{th}$ element of the vector $v$, and we define $\texttt{diag}(v)$ as a diagonal matrix whose $i^{th}$ diagonal element is $v_i$. A block diagonal matrix $\texttt{blkdiag}(A,B) = \begin{bmatrix} A & 0 \\ 0 & B \end{bmatrix}$ is also defined for matrices $A,B$. As shorthands, we use $c(\cdot)$, $s(\cdot)$ to denote $\cos(\cdot)$ and $\sin(\cdot)$. For vectors $v,w$, $[v;w] := [v^\top w^\top]^\top$. Column vectors of size $n$ whose elements are all $1$ and are all $0$ are denoted as $\bm{1}_n$ and $\bm{0}_n$, respectively. The identity matrix is defined as $I_n \in \mathbb{R}^{n \times n}$. Lastly, for a vector $v\in \mathbb{R}^n$, $\Tanh(v) := [\tanh(v_1);\tanh(v_2);\cdots;\tanh(v_n)]$ and $\Cosh(v) := \texttt{diag}([\cosh(v_1);\cosh(v_2);\cdots;\cosh(v_n)])$.

We also list properties related to $\tanh$ as follows \cite{fischer2014saturated}:
\begin{equation} \label{eq: tanh property}
\begin{gathered}
    \norm{x}^2 \geq \sum_{i=1}^n \text{ln}(\text{cosh}(x_i)) \geq \frac{1}{2} \text{tanh}^2(\norm{x}) \\
    \norm{x} \geq \norm{\Tanh(x)}, \quad \norm{\Tanh(x)}^2 \geq \text{tanh}^2(\norm{x}).
\end{gathered}
\end{equation}

\section{Equations of Motion}

In defining the equations of motion (EoM) of a fully actuated multirotor, we consider a single rigid body dynamics. We denote position, orientation, and body angular velocity of the multirotor as $p, \phi, \omega \in \mathbb{R}^3$, respectively. We use Euler angles $\phi$ to represent the orientation. Inerital parameters are mass $m \in \mathbb{R}_{>0}$ and mass moment of inertia $J \in \mathbb{R}^{3\times 3}_{>0}$. The body angular velocity $\omega$ and the time derivative of the Euler angles $\dot{\phi}$ satisfy the following relationship $\omega = Q \dot{\phi}$ where $Q \in \mathbb{R}^{3 \times 3}$ is a corresponding Jacobian matrix. We describe the gravitational acceleration as $g \in \mathbb{R}_{>0}$ and $b_3 = [0;0;1]$.

\begin{figure}
    \centering
    \includegraphics[width=0.65\linewidth]{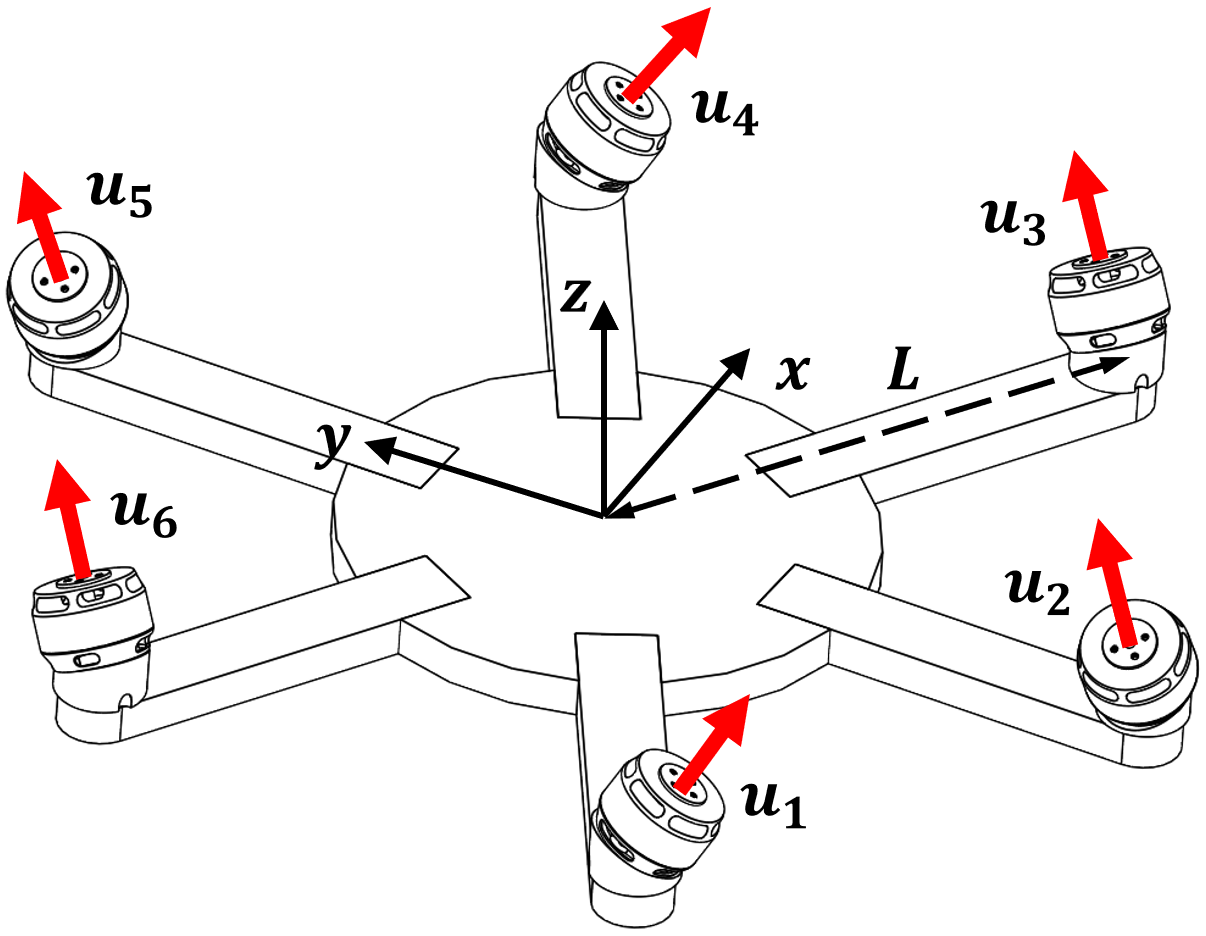}
    \caption{A fully actuated multirotor considered in this work. $u = [u_1, \cdots, u_6]^\top$ indicates the rotor thrust, and $L$ is the distance from the geometric center of the multirotor to a rotor.}
    \label{fig:multirotor}
\end{figure}

Then, the equations of motion of a fully actuated multirotor can be written as
\begin{equation} \label{eq; equation of motion}
\begin{aligned}
    m \ddot{p} &= R f - mgb_3 + d_t \\
    JQ \ddot{\phi} &= \tau - J \dot{Q} \dot{\phi} - \omega^\wedge J \omega + d_r
\end{aligned}
\end{equation}
where $d_t, d_r \in \mathbb{R}^3$ are external disturbance in translational and rotation motions, respectively. To allow designing a controller for the rotor thrust so that rotor thrust saturation can be directly taken into account, we find a relationship between the body force and torque $[f;\tau] \in \mathbb{R}^6$ and the rotor thrust $u = [u_1;\cdots;u_6] \in \mathbb{R}^6$. Using the geometric configuration of a fully actuated multirotor in Fig. \ref{fig:multirotor}, the relationship can be written as $[f;\tau] = Au$ where
\begin{equation} \label{eq: matrix A}
    A = \left[ \begin{matrix} 
    -\frac{1}{2}s\alpha & -\frac{1}{2}s\alpha & s\alpha & -\frac{1}{2}s\alpha & -\frac{1}{2}s\alpha & s\alpha \\
    -\frac{\sqrt{3}}{2}s\alpha & \frac{\sqrt{3}}{2}s\alpha & 0 & -\frac{\sqrt{3}}{2}s\alpha & \frac{\sqrt{3}}{2}s\alpha & 0 \\
    c\alpha & c\alpha & c\alpha & c\alpha & c\alpha & c\alpha \\
    -\frac{1}{2}P_1 & \frac{1}{2}P_1 & P_1 & \frac{1}{2}P_1 & -\frac{1}{2}P_1 & -P_1 \\
    -\frac{\sqrt{3}}{2}P_1 & -\frac{\sqrt{3}}{2}P_1 & 0 & \frac{\sqrt{3}}{2}P_1 & \frac{\sqrt{3}}{2}P_1 & 0 \\
    -P_2 & P_2 & -P_2 & P_2 & -P_2 & P_2
    \end{matrix} \right].
\end{equation}
Here, $P_1 = Lc\alpha - k_f s\alpha$, $P_2 = Ls\alpha+k_f c\alpha$ and $L, \alpha, k_f \in \mathbb{R}_{>0}$ are the length from the multirotor origin to a rotor, fixed tilt angle of a rotor, and a thrust-to-torque coefficient.

Then, the EoM can be compactly written in the following form:
\begin{equation} \label{eq; equation of motion - modified}
    M_n(\phi) \begin{bmatrix} \ddot{p} \\ \ddot{\phi} \end{bmatrix} 
    = A u + h_n(\phi,\dot{\phi}) + \begin{bmatrix} d_t \\ d_r \end{bmatrix} 
\end{equation}
where
\begin{equation*}
    M_n(\phi) = \left[\begin{matrix} m R^\top & 0 \\ 0 & JQ \end{matrix} \right], \ h_n(\phi,\dot{\phi}) = \left[\begin{matrix} -mgR^\top b_3 \\ -J\dot{Q}\dot{\phi} - \omega \times J\omega \end{matrix} \right].
\end{equation*}

\section{Controller Design}


\subsection{EoM reformulation}
To facilitate controller design and stability analysis, we reformulate EoM (\ref{eq; equation of motion - modified}) to have 1) symmetric mass matrix and 2) symmetric input bound (i.e. input bound of the form $-\overline{u} \leq u \leq \overline{u}$). First, we ensure symmetricity of the mass matrix by left-multiplying $G = \texttt{blkdiag}(R,Q^\top)$ to (\ref{eq; equation of motion - modified}). Next, we define a new control input $v = u - u_m$ that satisfies $-\overline{v} \leq v \leq \overline{v}$ where $u_m = (\overline{u} + \underline{u})/2$ and $\overline{v} = (\overline{u} - \underline{u})/2$. The reformulated EoM can be arranged for $q = [p;\phi]$ as
\begin{equation} \label{eq: equation of motion - final}
    M(q) \ddot{q} = G(q) A v + h(q,\dot{q}) + d(t)
\end{equation}
where $M(q) = \texttt{blkdiag}(m I_3, Q^\top J Q)$, $h(q,\dot{q}) = G(q) (h_n(q,\dot{q}) + A u_m)$, and $d = G(q) [d_t;d_r]$. Assuming that the orientation is confined in $S = \{q | \lvert \phi_1 \rvert, \lvert \phi_2 \rvert < \rho \}$ where $\rho < \pi/2$, $Q$ is nonsingular. Thus, from the fact that $m$ being a positive constant and $J$ being a positive definite constant matrix, the mass matrix satisfies the following property for some $\overline{m},\underline{m} > 0$:
\begin{equation} \label{eq: mass matrix property}
    \underline{m} \leq \lVert M(q) \rVert \leq \overline{m} \quad \forall q \in S.
\end{equation}

\subsection{Saturated robust control}
Considering system dynamics (\ref{eq: equation of motion - final}) and the control input $v$, we aim to design a controller that satisfies the following conditions: 1) $-\overline{v} \leq v \leq \overline{v}$ and 2) $q \rightarrow q_d$.
Accordingly, we design a saturated robust controller motivated by \cite{fischer2014saturated,fischer2012saturated} in which a control law for $u$ is designed for a dynamics of the following structure:
\begin{equation} \label{eq: dynamics in the previous paper}
    M(q) \ddot{q} = f(q,\dot{q},t) + u + d(t).
\end{equation}
Compared to the above dynamics (\ref{eq: dynamics in the previous paper}) considered in the previous papers, dynamics considered in this work (\ref{eq: equation of motion - final}) additionally includes a non-diagonal, configuration-dependent input matrix $G(q) A$. Although the same control law in the previous papers can be employed by introducing a new control input and finding a conservative, symmetric input bound for the new input, such method can excessively shrink the allowable control input\footnote{As done in \cite{fischer2014saturated}, if a new control input $\mu = G(q) A v$ exists whose symmetric bound $\overline{\mu} > 0$ satisfies $A^{-1} G(q)^{-1} \overline{\mu} \leq \overline{v}$ for all $q$, then the same control law in the previous papers can be utilized with the newly defined control input $\mu$. However, such $\overline{\mu}$ may not exist or result in an excessively small bound, especially considering a non-diagonal matrix $A$.}. Thus, in this section, we propose a modified control law that can additionally consider a time-varying input matrix while not sacrificing the input bound.

Before presenting a controller, we define error variables $e_1, e_2, e_f$ as
\begin{equation} \label{eq: error variables}
\begin{aligned}
    e_1 &= q_d - q \\
    e_2 &= \dot{e}_1 + \Lambda_1 \Tanh(e_1) + e_f \\
    \dot{e}_f &= -\Gamma_1 e_2 + \Tanh(e_1) - \Gamma_2 e_f
\end{aligned}
\end{equation}
where $\Lambda_1, \Gamma_1, \Gamma_2 \in \mathbb{R}^{n \times n}_{>0}$ are diagonal gain matrices. From below, if not ambiguous, we omit any dependence of a variable (e.g. $M(q) \text{ to } M$) for brevity. Based on these filtered error variables, we construct the following saturated robust controller:
\begin{equation} \label{eq: rise with saturation}
\begin{aligned}
    v &= & & \Gamma_1 \Tanh(z) \\
    \dot{z} &= & &\Cosh^2(z) \Gamma_1^{-1} A^{-1} G^{-1} \Big( M \Gamma_1 \big\{ \Lambda_2 \Tanh(e_2) + \\
    & & &\Lambda_3 e_2 + \Gamma_2 e_2 \big\} + \Theta \sgn(e_2) - \dot{G}Av \Big)
\end{aligned}
\end{equation}
where $\Lambda_2, \Lambda_3, \Theta \in \mathbb{R}^{n \times n}_{>0}$ are also diagonal gain matrices. By taking $\Gamma_1 = \texttt{diag}(\overline{v})$, it can be easily observed that the input bound condition $-\overline{v} \leq v \leq \overline{v}$ is satisfied. 

\section{Stability Analysis}
To analyze stability of the closed-loop system composed of (\ref{eq: equation of motion - final}) and (\ref{eq: rise with saturation}), we first define a variable $r$ to ease the analysis:
\begin{equation*}
    r = \dot{e}_2 + \Lambda_2 \Tanh(e_2) + \Lambda_3 e_2.
\end{equation*}
Using (\ref{eq: equation of motion - final}), (\ref{eq: error variables}), and the definition of $r$, the following can be established.
\begin{equation*}
\begin{aligned}
    M r &= & &S - h_d + M \ddot{q}_d - d - GAv - M \Gamma_1 e_2 \\
    S &= & &M \Lambda_1 \Cosh^{-2}(e_1) \big( e_2 - \Lambda_1 \Tanh(e_1) - e_f \big) + \\
    & & & M \left( \Tanh(e_1) - \Gamma_2 e_f \right) + M \Lambda_2 \Tanh(e_2) + \\
    & & & M \Lambda_3 e_2 + (h_d - h)
\end{aligned}
\end{equation*}
where $h_d = h(q_d,\dot{q}_d)$. Then, we obtain the following using the time derivative of the above equation and (\ref{eq: rise with saturation}).
\begin{equation} \label{eq: M rdot}
\begin{aligned}
    M \dot{r} &= & &\tilde{N} + N_d - M \Gamma_1 r - \Theta \sgn(e_2) - \tfrac{1}{2} \dot{M} r - \\
    & & & \Tanh(e_2) - e_2 \\
    \tilde{N} &= & &\dot{S} - M \Gamma_1 \Gamma_2 e_2 + \Tanh(e_2) + e_2 - \dot{M}(\Gamma_1 e_2 + \tfrac{1}{2} r) \\
    N_d &= & & \dot{M} \ddot{q}_d + M \dddot{q}_d - \dot{h}_d - \dot{d}
\end{aligned}
\end{equation}
For the analysis, we assume that there exist $\zeta_{N_{d1}}, \zeta_{N_{d2}} \in \mathbb{R}^n_{>0}$ satisfying $\zeta_{N_{d1},i} = \sup_t \lvert N_{d,i}(t) \rvert$ and $\zeta_{N_{d2},i} = \sup_t \lvert \dot{N}_{d,i}(t) \rvert$.


Now, we take a candidate Lyapunov function $V$ as
\begin{equation} \label{eq: Lyapunov function}
\begin{aligned}
    V &= & & \sum^n_{i=1} \text{ln}(\Cosh(e_{1i})) + \sum^n_{i=1} \text{ln}(\Cosh(e_{2i})) + \\
    & & & \cfrac{1}{2} e_2^\top e_2 + \cfrac{1}{2} r^\top M r + \cfrac{1}{2} e_f^\top e_f + P
\end{aligned}
\end{equation}
where $P$ satisfies the following:
\begin{equation} \label{eq: P conditions}
\begin{aligned}
    \dot{P} &= & & - r^\top (N_d - \Theta \sgn(e_2)) \\
    P(t_0) &= & & \sum^n_{i=1} \theta_i \lvert e_{2i}(t_0) \rvert - e_2(t_0)^\top N_d(t_0).
\end{aligned}
\end{equation}
\begin{lemma}
By taking a sufficient large $\Theta$ satisfying $\theta_i > \zeta_{N_{d1},i} + \lambda_{3,i}^{-1} \zeta_{N_{d2},i}$, then $P(t) \geq 0$ $\forall t \geq t_0$.
\end{lemma}
\begin{proof}
    The proof is finished using the diagonal property of $\Lambda_2, \Lambda_3, \Theta$ and \cite[Lemma 3]{fischer2012saturated}.
\end{proof}

\begin{theorem}
    Assuming that control gains satisfy the following:
    \begin{equation*}
    \begin{gathered}
        \underline{\lambda}_1 > \cfrac{1}{2}, \quad \underline{\lambda}_3 > \cfrac{1}{2} + \cfrac{\bar{\gamma}_1^2 \xi^2}{4}, \quad \underline{\gamma}_2 > \cfrac{1}{\xi^2}, \quad \underline{m} \underline{\gamma}_1 > \eta \\
         \theta_i > \zeta_{N_{d1},i} + \lambda_{3,i}^{-1} \zeta_{N_{d2},i} \quad \forall i, \quad 4 \eta \mu > \rho^2(\norm{w(0)})
    \end{gathered}
    \end{equation*}
    where $\eta, \xi \in \mathbb{R}_{>0}$ are adjustable positive constants to be defined, and $\underline{a}, \overline{a} \in \mathbb{R}$ are minimum and maximum eigenvalues of a matrix $A$. A concatenated error vector is defined as $w = [\Tanh(e_1); e_2; r; e_f]$. Then, the closed-loop system consisting of dynamics (\ref{eq: equation of motion - final}) and controller (\ref{eq: rise with saturation}) is locally asymptotically stable.
\end{theorem}
\begin{proof}   
    Using the symmetric property of $M$ matrix, (\ref{eq: tanh property}), (\ref{eq: M rdot}), and (\ref{eq: P conditions}), we can compute the following inequality for $\dot{V}$.
\begin{equation*}
\begin{aligned}
    \dot{V} &= & & \dot{e}_1^\top \Tanh(e_1) + \dot{e}_2^\top \Tanh(e_2) + \dot{e}_2^\top e_2 + \dot{r}^\top M r + \\
    & & &\cfrac{1}{2} r^\top \dot{M} r + \dot{e}_f^\top e_f + \dot{P} \\
    &\leq & & -\underline{\lambda}_1 \norm{\Tanh(e_1)}^2 - (2 \underline{\lambda}_2 + \underline{\lambda}_3) \norm{\Tanh(e_2)}^2 - \\
    & & & \underline{\lambda}_3 \norm{e_2}^2 - \underline{\gamma}_2 \norm{e_f}^2 - \underline{m}\underline{\gamma}_1 \norm{r}^2 + r^\top \tilde{N} -\\
    & & & e_f^\top \Gamma_1 e_2 + \Tanh(e_1)^\top e_2
\end{aligned}
\end{equation*}
The upper bound of $\tilde{N}$ can be expressed as $\norm{\tilde{N}} \leq \rho(\norm{w}) \norm{w}$ using the Mean Value Theorem. $\rho(\cdot) \in \mathbb{R}$ is a positive, strictly increasing function.

We further elaborate $\dot{V}$ as
\begin{equation} \label{eq: Vdot}
    \dot{V} \leq -\left(\mu - \cfrac{\rho^2(\norm{w})}{4 \eta} \right) \norm{w}^2
\end{equation}
where 
\begin{equation*}
    \mu = \min \left\{ \underline{\lambda}_1 -\cfrac{1}{2}, 
    \underline{\lambda}_3 - \cfrac{1}{2} -\cfrac{\bar{\gamma}_1^2 \xi^2}{4},
    \underline{\gamma}_2 - \cfrac{1}{\xi^2},
    \underline{m} \underline{\gamma}_1 - \eta
    \right\}.
\end{equation*}
 In the derivation, the following Young's inequalities are employed for $\eta, \xi > 0$:
\begin{equation*}
\begin{aligned}
    \rho(\norm{w})\norm{w}\norm{r} &\leq \cfrac{\rho^2(\norm{w})}{4 \eta} \norm{w}^2 + \eta \norm{r}^2 \\
    \bar{\gamma}_1 \norm{e_f} \norm{e_2} &\leq \cfrac{1}{\xi^2} \norm{e_f}^2 + \cfrac{\bar{\gamma}_1^2 \xi^2}{4} \norm{e_2}^2.
\end{aligned}
\end{equation*}

Let $x = [e_1; e_2; r; e_f]$, $y = [x; \sqrt{P}]$, and define a domain $\mathcal{D} = \{y\in \mathbb{R}^{4n+1} | \norm{y} \leq \rho^{-1}(2\sqrt{\mu \eta}) \}$. Using (\ref{eq: tanh property}) and Lemma 1, the lower and upper bounds of $V$ can be obtained as
\begin{equation} \label{eq: V bound}
    \phi_1(y) \leq V(y) \leq \phi_2(y) \quad \forall y \in \mathcal{D}
\end{equation}
where $\phi_1(y) = \tfrac{1}{2} \min (1,\underline{m}) \text{tanh}^2(\norm{y})$ and $\phi_2(y) = \frac{1}{2} \max (\tfrac{1}{2}\overline{m},\tfrac{3}{2}) \norm{y}^2$. Revisiting the definition of $\mathcal{D}$ and (\ref{eq: Vdot}), the following holds for $\phi(y) = c \text{tanh}^2(\norm{x})$ and some positive constant $c$.
\begin{equation} \label{eq: V upper bound}
    \dot{V} \leq -\phi(y) \quad \forall y \in \mathcal{D}.
\end{equation}
Now, asymptotic stability can be derived invoking \cite[Corollary 1]{fischer2013lasalle}.
\end{proof}
    
\section{Results}

\begin{table} 
\centering
\caption{Simulation settings}
\label{tb:simulation settings}
\begin{tabular}{@{}c c c @{}} 
\toprule
Parameter & Value & Unit \\ 
\midrule
    \begin{tabular}{@{}l@{}} $q(0)$ \\ $q_d(t)$  \\ $d(t)$ \\ $L$ \\ $k_f$ \\ $\alpha$ \\ $\overline{u}$ \\ $\underline{u}$ \\ $m$ \\ $J_b$ \end{tabular} 
&
    \begin{tabular}{@{}l@{}} $[0;0;0;0;0;0]$ \\ $[\cos(\tfrac{\pi}{5} t); \sin(\tfrac{\pi}{5} t); 1.0; 0;0;0]$ \\ $[5 \sin (\tfrac{\pi}{5} t); 0; -5; 0; 0.05; 0]$ \\ $0.258$ \\ $0.016$ \\ $30$ \\ $20 \bm{1}_6$ \\ $\bm{0}_6$ \\ $2.9$ \\ $\texttt{diag}([0.035; 0.035; 0.045])$ \end{tabular}
&
    \begin{tabular}{@{}l@{}} \si{m} or \si{rad} \\ \si{m} or \si{rad}  \\ \si{N} or \si{Nm} \\ \si{m} \\ \si{m} \\ \si{deg} \\ \si{N} \\ \si{N} \\ \si{kg} \\ \si{kg m^2} \end{tabular} \\
\bottomrule
\end{tabular}
\end{table}

\begin{table} 
\centering
\caption{Controller parameters}
\label{tb:parameters/gains}
\begin{tabular}{@{}l l | l l@{}} 
\toprule
Parameter & Value  & Parameter & Value\\ 
\midrule
    \begin{tabular}{@{}l@{}} $\Gamma_1$ \\ $\Gamma_2$ \\ $\Theta$ \end{tabular} 
&
    \begin{tabular}{@{}l@{}} $10 I_6$ \\ $I_6$ \\ $\texttt{blkdiag}(20 I_3, 0.1 I_3)$ \end{tabular}
&
    \begin{tabular}{@{}l@{}} $\Lambda_1$ \\ $\Lambda_2$ \\ $\Lambda_3$ \end{tabular}
& 
    \begin{tabular}{@{}l@{}} $2 I_6$ \\ $10 I_6$ \\ $10 I_6$ \end{tabular} \\
\bottomrule
\end{tabular}\\
\end{table}

\subsection{Simulation setup}
We validate the proposed controller in simulation. The simulation scenario is tracking a circular trajectory in the presence of time-varying disturbance. We devise this scenario to illustrate that the proposed controller produces a control input that abides to the rotor thrust limit while inducing asymptotic stability. Detailed settings regarding initial condition, ineritial parameters of the multirotor, disturbance, and minimum and maximum rotor thrust limits are listed in Table \ref{tb:simulation settings}. Parameters $L,k_f,\alpha$ related to the matrix $A$ in (\ref{eq: matrix A}) are also presented in the Table. Table \ref{tb:parameters/gains} shows the selected controller gains. In selecting the control gains, we first set $\Gamma_1 = \texttt{diag}(\overline{u} - \underline{u})/2 = 10 I_6$. Then, $\Lambda_2, \Lambda_3$ are set to the same value, and we tune them starting from $I_6$. Considering the controller structure (\ref{eq: rise with saturation}), we set $\Theta$ to a smaller scale compared to $M \Gamma_1 \Lambda_2$. Although the term $\Theta \sgn(e_2)$ may trigger input chattering, the term plays a key role in achieving asymptotic stability (as illustrated in Figs. \ref{fig:result_woRISE}, \ref{fig:result_wRISE} and \ref{fig:result_xyPlane}). We tune $\Theta$ while taking this trade-off into account. $\Lambda_1, \Gamma_2$ are tuned by first setting them as $I_6$.

We conduct three different simulations. In the first simulation, we perform comparative study using a method in \cite{fischer2012saturated, fischer2014saturated} with suitable modifications where no state-dependent input matrix is allowed. In the second simulation, we set $\Theta$ to zero so that RISE (robust integral of the sign of the error) does not contribute to the control input. Finally, the last simulation is performed with a non-zero $\Theta$ whose value is in Table \ref{tb:parameters/gains}. By comparing the first and last simulation results, we validate effectiveness of the proposed method in handling the non-diagonal, state-dependent input matrix. Furthermore, we illustrate the strength of the $\sgn$ term in achieving asymptotic stability by comparing the second and third simulation results. Simulation results are visualized in Figs. \ref{fig:result_woProposed}, \ref{fig:result_woRISE}, \ref{fig:result_wRISE}, and \ref{fig:result_xyPlane}. 


In implementing \cite{fischer2012saturated,fischer2014saturated}, since a state-dependent input matrix is not allowed, we define a virtual control input $v_c = G(q) A v$ by which the EoM (\ref{eq: equation of motion - final}) can be written as 
\begin{equation*}
    M(q) \ddot{q} = v_c + h(q,\dot{q}) + d(t).
\end{equation*}
The input bound for $v_c$ then needs to be defined such that for every $v_c \in [-\overline{v}_c, \overline{v}_c]$, $v = A^{-1} G(q)^{-1} v_c$ satisfies $v \in [-\overline{v}, \overline{v}]$. As done in \cite{fischer2014saturated}, using the fact that $\lVert A b \rVert_\infty \leq \lVert A \rVert_\infty \lVert b \rVert_\infty$ for every matrix $A \in \mathbb{R}^{n \times n}$ and vector $b \in \mathbb{R}^n$, we select this conservative bound $\overline{v}_c$ as
\begin{equation} \label{eq: input bound comparison}
    \overline{v}_c = \frac{\min \overline{v}_i}{\lVert A^{-1} G(0)^{-1} \rVert_\infty} \bm{1}_n
\end{equation}
where we impose a small angle assumption for roll and pitch angles, leading to $G(q) \approx I_n$. Thus, from the definition of $A$ in (\ref{eq: matrix A}) and parameters in Table \ref{tb:simulation settings}, we compute the above bound (\ref{eq: input bound comparison}) and set $\overline{v}_c = 2.28 \bm{1}_n$ during simulation.

Then, the control law can be written as 
\begin{equation*}
\begin{aligned}
    v_c &= & & \Gamma_1 \Tanh(z_c) \\
    \dot{z_c} &=  & &\Cosh^2(z_c) \Gamma_1^{-1} \Big( M \Gamma_1 \big\{ \Lambda_2 \Tanh(e_2) + \Lambda_3 e_2 + \Gamma_2 e_2 \big\} + \\
    & & & \Theta \sgn(e_2)\Big)
\end{aligned}
\end{equation*}
where $\Gamma_1$ here is $\Gamma_1 = \texttt{diag}(\overline{v}_c)$, and the other gains follow the values in Table \ref{tb:parameters/gains}.

\subsection{Analysis}
\begin{figure}
    \centering
    \includegraphics[width=1.0\linewidth]{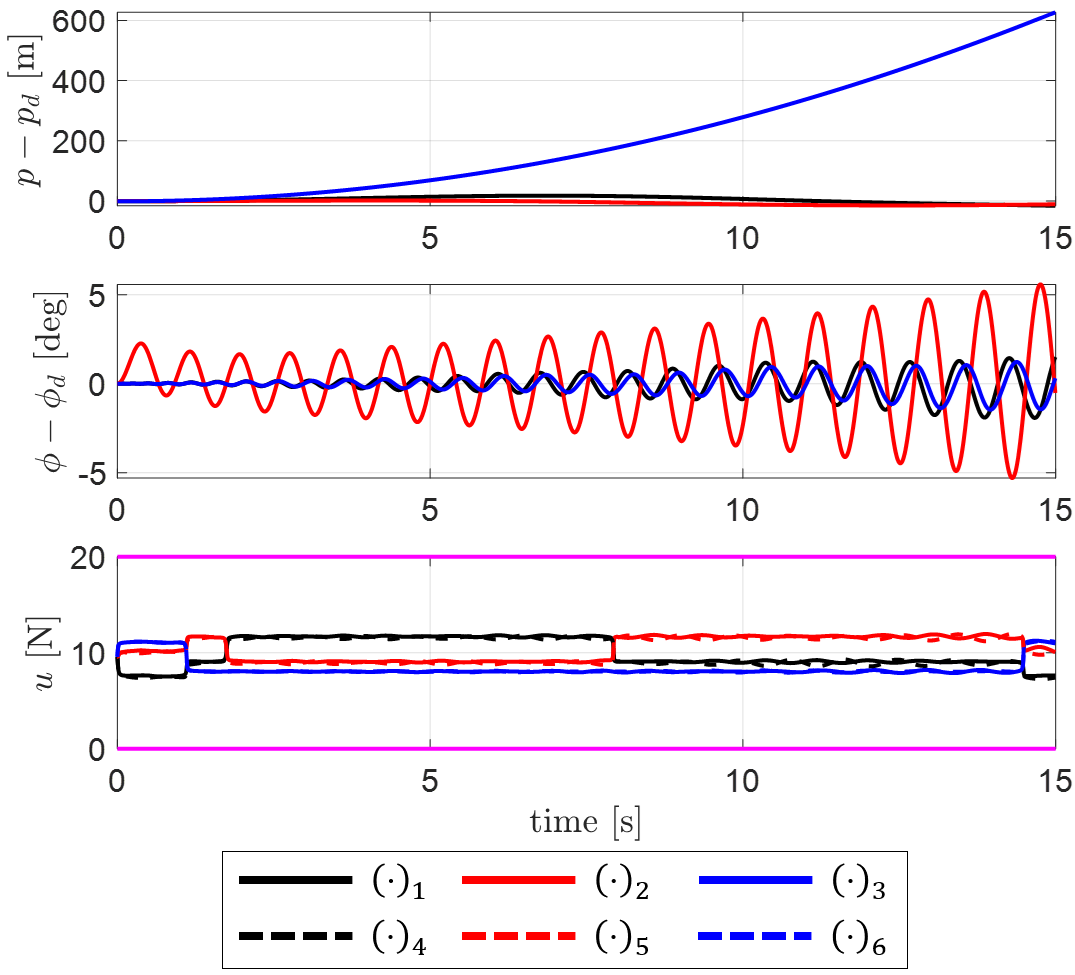}
    \caption{Simulation 1 -- results of modified-\cite{fischer2012saturated,fischer2014saturated}.}
    \label{fig:result_woProposed}
\end{figure}

\begin{figure}
    \centering
    \includegraphics[width=1.0\linewidth]{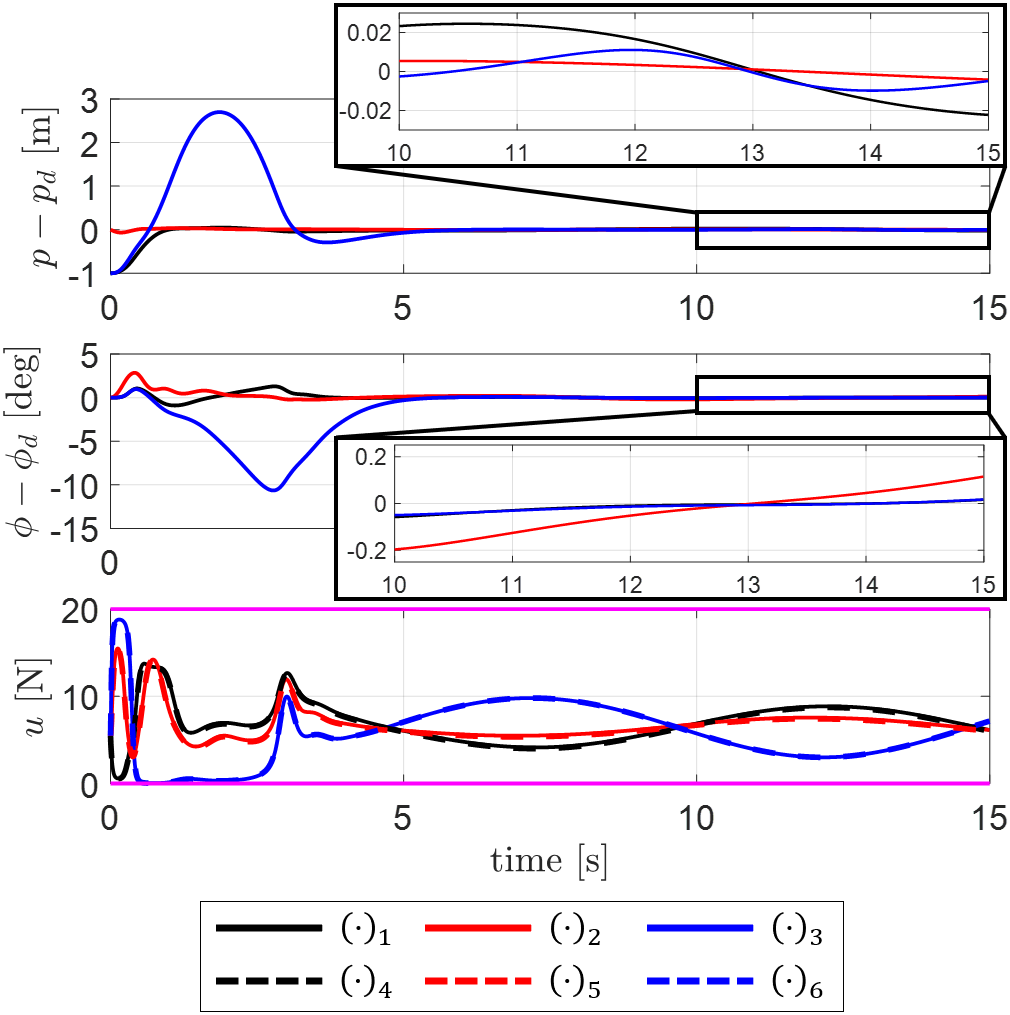}
    \caption{Simulation 2 -- results without the $\sgn$ term.}
    \label{fig:result_woRISE}
\end{figure}

\begin{figure}
    \centering
    \includegraphics[width=1.0\linewidth]{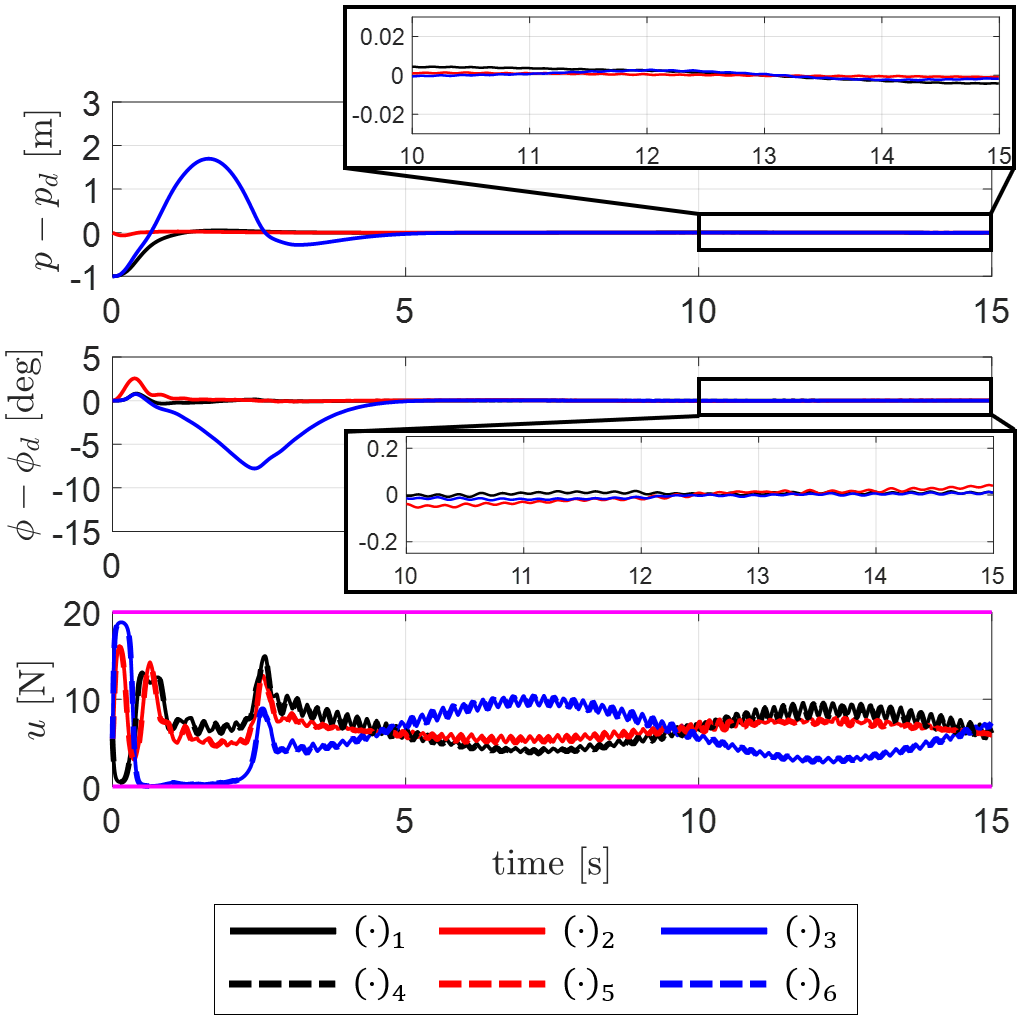}
    \caption{Simulation 3 -- results with the proposed method.}
    \label{fig:result_wRISE}
\end{figure}

For the first simulation result in Fig. \ref{fig:result_woProposed}, errors in both translation and rotation directions diverge. This is mainly due to the conservative input bound which is resulted from inability of a controller to handle a non-diagonal, state-dependent input matrix. The shrunk input bound compared to the actual input bound $\overline{u}, \underline{u}$ can be observed in the bottom graph in Fig. \ref{fig:result_woProposed}. On the other hand, by incorporating the input matrix structure in the controller, the actual input bound can be fully exploited in the proposed controller. Accordingly, as in Fig. \ref{fig:result_wRISE}, stability is preserved during the same simulation setting.

Next, we compare the results from simulation 2 and 3. As can be observed in the bottom graphs of Figs. \ref{fig:result_woRISE} and \ref{fig:result_wRISE}, the control inputs are bounded between the maximum and minimum values marked in magenta solid lines. Such property is endowed by the structure of the proposed control law where rotor thrust saturation is directly addressed using the $\tanh$ function. Input chattering can be observed in the bottom graph of Fig. \ref{fig:result_wRISE} as in consequence of introducing $\sgn$ while Fig. \ref{fig:result_woRISE} shows no such phenomenon. As will be discussed below, this is a trade-off for achieving asymptotic stability. To mitigate this issue, one possible approach would be replacing $\sgn$ with a continuous counterpart.

\begin{figure}
    \centering
    \includegraphics[width=0.8\linewidth]{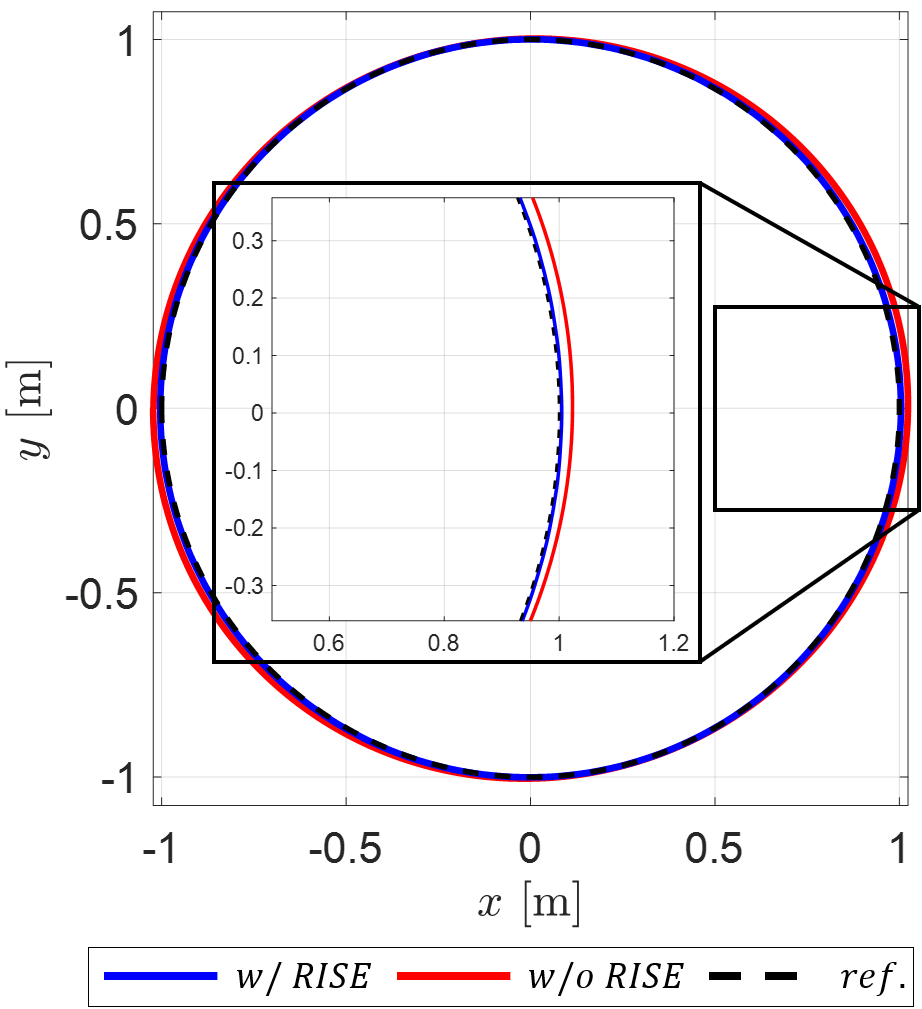}
    \caption{Comparison between the results of simulation 2 and 3. To compare the results during the steady-state phase, we only plot the results after the transient phase (i.e. $t \geq 5$ \si{s}).}
    \label{fig:result_xyPlane}
\end{figure}

By comparing the results of Figs. \ref{fig:result_woRISE} and \ref{fig:result_wRISE}, particularly the data in the enlarged graphs, it can be confirmed that $\sgn$ function does contribute to asymptotic stability. The result with non-zero $\Theta$ in Fig. \ref{fig:result_wRISE} shows much smaller magnitude of errors in the enlarged graphs than the one with $\Theta=0$ in Fig. \ref{fig:result_woRISE}. We further visualize this analysis in Fig. \ref{fig:result_xyPlane} where the results of the two simulations are projected to the XY plane. Since this figure is to analyze asymptotic behavior, only the results after the transient phase, i.e. $t \geq 5$ \si{s}, are visualized. The trajectory obtained from the proposed method (blue solid line) outperforms that without the $\sgn$ term (red solid line) in tracking the reference trajectory displayed with the black dashed line.  

\section{Conclusion}
This work presents a saturated RISE control for a fully actuated multirotor to consider rotor thrust saturation and disturbance rejection. We first reformulate the dynamics of a fully actuated multirotor to have a symmetric input bound and a symmetric mass matrix. Then, compared to the existing saturated RISE controller, we modify the control law to be capable of handling a non-diagonal, state-dependent input matrix. By directly incorporating the input matrix structure in the control law, the control law fully exploits the whole input bound unlike the previous works which sacrifice the allowable input bound. We present formal stability analysis for local asymptotic stability even under time-varying external disturbance. The proposed method is validated in simulation where we conduct comparative studies with the previous work which requires input bound shrinking, and with the case without the $\sgn$ term in the control law. As a future work, we aim to perform hardware experiments.

\addtolength{\textheight}{-12cm}   

\normalem 


\end{document}